# Interpretable Synthetic Medical Tabular Data Generation for Clinical Decision Support Using Fuzzy Cognitive Maps

Michael Vasilakakis
Department of Computer Science and Biomedical Informatics
University of Thessaly
Lamia, Greece
vasilaka@uth.gr

Dimitris K. Iakovidis
Department of Computer Science and Biomedical Informatics
University of Thessaly
Lamia, Greece
diakovidis@uth.gr

***Abstract*— Synthetic medical tabular data generation has become essential for developing and validating computer-based medical systems (CBMSs) when real clinical data is restricted due to privacy, ethical, or data availability limitations. Existing probabilistic and deep generative models often lack interpretability and fail to preserve clinically meaningful dependencies, limiting their suitability for safety-critical applications. This paper proposes a novel application of Fuzzy Cognitive Maps (FCMs) in a framework for synthetic medical tabular data generation with explicit causality and privacy preservation. Clinical features are described using linguistically interpretable fuzzy sets, and inter-feature dependencies are encoded as FCM edge weights computed from fuzzy set intersections. Synthetic patient records are generated by propagating randomly initialized linguistic activation vectors through the FCM until convergence, followed by defuzzification to produce clinically coherent numerical values. The approach natively handles mixed data types, and domain constraints common in health records. Experimental evaluation on UCI medical benchmark datasets demonstrates competitive performance under a Train-on-Synthetic-Test-on-Real (TSTR) protocol. The proposed method achieves accuracy of up to 0.81 and AUROC of up to 0.90 on the Heart Disease dataset, matching or exceeding TVAE and Gaussian Copula baselines while running exclusively on CPU. Fidelity metrics including KS Complement (up to 0.91) and Correlation Similarity (up to 0.95) confirm strong statistical coherence, and DCR Baseline Protection scores consistently exceed those of TVAE, confirming adequate privacy guarantees. These results demonstrate that causally grounded, interpretable fuzzy modeling offers a computationally efficient and transparent alternative to deep generative models for trustworthy synthetic data generation in CBMSs.***

## I. INTRODUCTION

Machine learning (ML) has become a core component of modern computer-based medical systems, enabling applications such as early disease detection, clinical risk stratification, personalized treatment planning, and public health surveillance. These systems rely on large volumes of representative data to achieve reliable performance. In healthcare, however, the availability of such data is frequently limited by privacy regulations, ethical constraints, and institutional policies governing the use of electronic health records. As a result, many promising ML-driven medical systems face challenges in development, validation, and deployment. Synthetic data generation has emerged as an effective strategy to mitigate these limitations by enabling data-driven modeling while reducing the exposure of sensitive patient information. Advances in generative modeling have significantly increased the realism and practical utility of synthetic data, making it a valuable asset for medical research and system development without violating regulatory frameworks [1]. Recent advances in data acquisition and digital infrastructure have enabled large-scale collection and analysis of clinical and biomedical data, supporting a transition from experience-driven practice to evidence-based decision making. While this transition has accelerated innovation in healthcare analytics, it has also intensified concerns related to data sharing and secondary data use. Clinical data often contain sensitive personal, institutional, and contextual information, and their dissemination raises issues related to patient privacy, intellectual property, and cybersecurity. These concerns are particularly acute when data must be shared across organizational or national boundaries to support collaborative research and the development of robust medical systems.

Privacy-preserving strategies have been extensively investigated to address these challenges. Traditional anonymization and privacy-preserving data publishing techniques aim to reduce re-identification risk but often degrade data utility. More recent approaches, including distributed privacy-preserving data mining and federated learning, seek to avoid centralized data sharing by aggregating model updates rather than raw data [2]. Although effective in certain scenarios, such approaches introduce additional system complexity, communication overhead, and constraints on model design. In contrast, synthetic data generation enables the dissemination of data-like artifacts that preserve essential statistical and structural characteristics of the original datasets while minimizing privacy risks. By replacing real datasets with synthetic counterparts, sensitive information can be protected while still enabling downstream analysis, benchmarking, and model training. When access to real clinical data is restricted or insufficient, synthetic data provide a practical alternative for developing and validating computer-

This work is part of the European project SEARCH, which is supported by the Innovative Health Initiative Joint Undertaking (IHI JU) under grant agreement No. 101172997. The JU receives support from the European Union's Horizon Europe research and innovation programme and COCIR, EFPIA, Europa Bio, MedTech Europe, Vaccines Europe, Medical Values GmbH, Corsano Health BV, Syntheticus AG, Maggioli SpA, Motilent Ltd, Ubitech Ltd, Hemex Benelux, Hellenic Healthcare Group, German Oncology Center, Byte Solutions Unlimited, AdaptIT GmbH. Views and opinions expressed are however those of the author(s) only and do not necessarily reflect those of the aforementioned parties. Neither of the Synthetic data generation methods can be broadly divided into neural and non-neural network-based approaches. Neural methods, particularly Generative Adversarial Networks (GANs) and Variational Autoencoders (VAEs) [2], have been widely adopted for tabular data synthesis. These methods typically rely on extensive preprocessing, such as encoding categorical variables and normalizing continuous features, to accommodate neural architectures [3]. While they can achieve high fidelity, their opaque internal representations and computational demands limit their suitability for safety-critical medical applications, where transparency and controllability are essential.

Non-neural network-based methods remain relevant due to their interpretability and stronger alignment with domain constraints. Copula-based models [4] for example, explicitly model dependencies among variables and have been applied to synthetic tabular data generation. However, parametric copulas often encounter scalability limitations in high-dimensional settings and may struggle to capture complex, context-dependent relationships present in clinical data[3]. Consequently, selecting an appropriate synthetic data generation method remains a nontrivial task. In biomedical applications, even subtle inconsistencies or implausible relationships in synthetic data can undermine trust and limit clinical applicability, highlighting the importance of methods that preserve meaningful dependencies [1].

Tabular medical data presents additional challenges due to their heterogeneity, encompassing continuous measurements, categorical attributes, and discrete clinical indicators, often with skewed or imbalanced distributions. Data quality issues such as missing values, incomplete records, and limited cohort sizes further complicate modeling efforts. Moreover, clinical datasets vary substantially in dimensionality, feature composition, and data semantics, making it unlikely that a single generative approach will perform consistently across all scenarios [5].

A defining characteristic of medical tabular data is the presence of structured dependencies and logical constraints among variables. Unlike image or text data, tabular features do not exhibit inherent spatial or sequential structure, and their relationships are often governed by domain knowledge rather than statistical correlation alone. Clinical variables are semantically grounded and represent real physiological or pathological quantities, such as laboratory values, vital signs, or diagnostic indicators. As a result, modeling interactions among features and preserving causal or logical relationships is essential to ensure the plausibility and safety of synthetic medical data. For example, certain combinations of attributes, such as biological sex and pregnancy status, are governed by strict clinical constraints. Failure to respect such dependencies can lead to unrealistic or misleading synthetic records, limiting their usefulness in computer-based medical systems [6].

Uncertainty, incompleteness, and semantic ambiguity are inherent characteristics of clinical data and pose significant challenges for the development of reliable computer-based medical systems. In clinical decision support settings, models must not only achieve high predictive performance but also provide transparent and interpretable reasoning pathways that allow clinicians to understand, assess, and, when necessary, intervene in system behavior. Representations based on interpretable concepts offer a practical mechanism for addressing these requirements by enabling human-understandable abstractions of clinical variables and their interactions [7].

Fuzzy Cognitive Maps (FCMs) provide a structured and intuitive modeling framework for representing such abstractions. By modeling systems as directed graphs of interpretable concepts connected through weighted causal influences, FCMs allow explicit encoding of clinically meaningful dependencies and constraints [8]. This causal structure supports reasoning about how changes in one clinical variable may influence others, facilitating scenario analysis and transparent decision logic. In biomedical contexts, where expert knowledge plays a critical role, FCMs further enable a human-in-the-loop paradigm, in which clinicians can inspect, validate, and refine the modeled relationships to ensure alignment with medical knowledge and practice [9].

Motivated by these properties, this paper introduces a synthetic medical tabular data generation framework that employs FCMs as a causal reasoning layer to ensure clinical coherence and interpretability. Rather than relying on opaque latent representations or purely statistical dependency modeling, the proposed approach generates synthetic patient records through concept-level reasoning, where clinical features are represented as interpretable states and their interactions are governed by an FCM-based influence structure. This design enables the generation process to be audited and adjusted by domain experts, supporting trust and transparency in safety-critical medical applications.

Synthetic records are produced by initializing and propagating concept activations through the FCM until a stable configuration is reached, after which clinically consistent feature values are realized. This process naturally supports mixed data types, missing values, and domain constraints commonly encountered in electronic health records, while preserving clinically meaningful dependencies. By maintaining an explicit causal and semantic structure throughout the generation process, the proposed framework facilitates explainable synthetic data generation that can be safely integrated into clinical decision support workflows, including model training, validation, and what-if analysis.

The remainder of this paper is organized as follows. Section II describes the proposed synthetic data generation framework and its underlying causal modeling components. Section III presents experimental results and comparisons with representative baseline methods on medical datasets. Section IV concludes the paper and discusses future research directions.

## II. FUZZY COGNITIVE MAPS

Fuzzy Cognitive Maps (FCMs) are powerful tools for modeling complex systems. An FCM is a graph-based, knowledge-driven model composed of nodes representing system concepts and weighted edges encoding the causal relationships among these concepts. An FCM is formally defined as a directed graph comprising $N$ concepts $C_i, i = 1,..,N$ and a set of weighted connections $w_{ji} \in [-1, 1]$, representing the strength and direction of influence from concept $C_i$ to concept $C_j$ [8]. Each of the $N$ concepts $C_i, i = 1, \dots, N$, of the FCM has a value $A_i \ \epsilon \ [0,1]$. Directed edges between concepts encode causal influence. There are three types of causal relationships: a) positive $w_{ji} > 0$ , indicating that an increase in the value of $C_j$ leads to an increase in the value of $C_i$, b) negative ($w_{ji} < 0$), indicating that increase in the value of $C_j$, causes a decrease of the value of $C_i$, and c) neutral $(w_{ji} = 0)$ , indicating the absence of a causal relationship between $C_j$ and $C_i$.

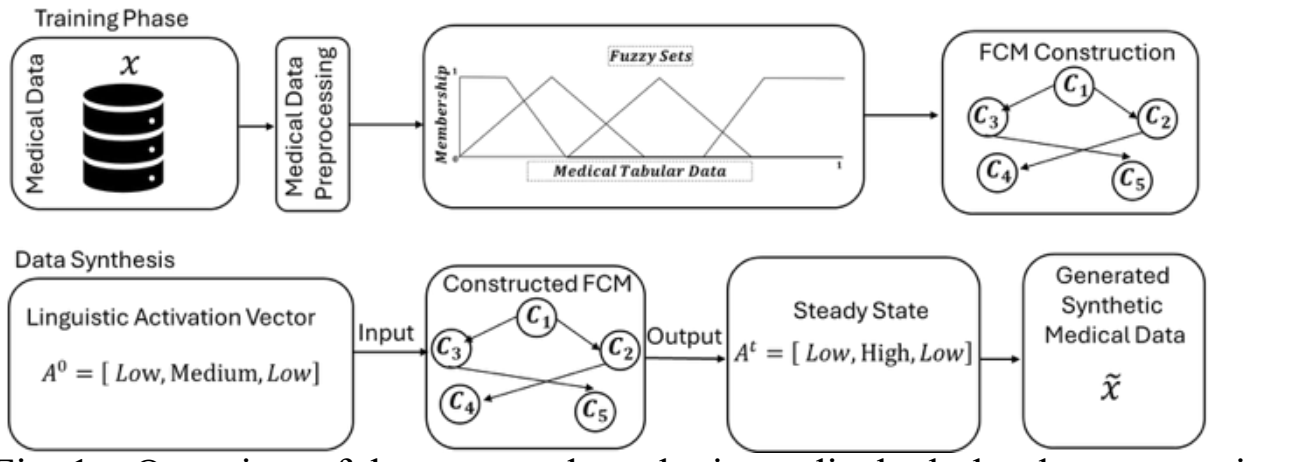

Fig. 1. Overview of the proposed synthetic medical tabular data generation based on FCM.

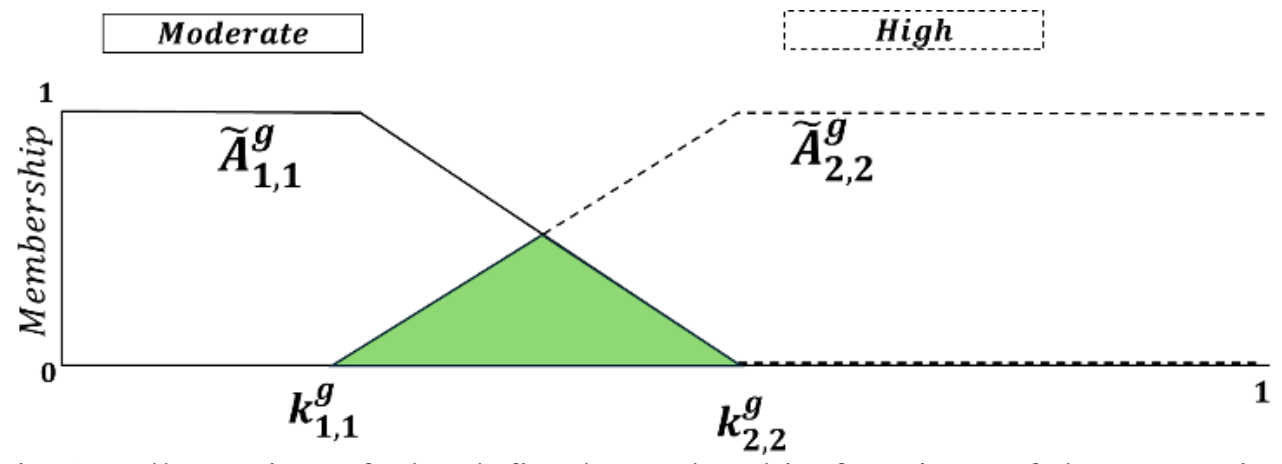

Fig. 2. Illustration of the defined membership functions of the respective fuzzy sets. Solid lines correspond to the features $k^g_{1,1}$, and the dashed lines correspond to the features $k^g_{2,2}$ . The green overlapping region highlights the intersection of the two membership functions, indicating the degree of simultaneous activation between the corresponding linguistic states.

Typically, for the definition of an FCM, domain experts are involved in the determination of concepts and their interconnections, utilizing fuzzy "IF-THEN" rules. Once the FCM structure is established, a reasoning process is initiated using a specific test case. This process proceeds iteratively until the system converges to a steady state. Convergence of the iterative update is not guaranteed in general, sufficient conditions based on the contraction mapping theorem for sigmoid-based FCMs are discussed in [10]. The activation values of the output concepts at this steady state are interpreted as the inferred outcomes of the modeled system. During this reasoning process, the concept activation values $A_i \in [0,1]$ are updated iteratively according to the following formulation:

$$A_i^{t+1} = f\left(A_i^t + \sum_{j=1, j\neq i}^{N} A_j^t \cdot w_{ji}\right) \quad (1)$$

where $A_i^{t+1}$ represents the value of $C_i$ at the iteration $t+1$, $w_{ji}$ is the influence of $C_j$ on $C_i$, and $f$ is a sigmoid function such as the log sigmoid, which maps the concept values within [0,1][11].

$$f(x) = \frac{1}{1+\exp(-x)} \quad (2)$$

The initial state vector $A^0$ represents the initial concept values, for $t = 0$.

## III. Medical Data Generation Framework

The proposed methodology addresses synthetic medical tabular data generation as a clinically interpretable and causally coherent reasoning process rather than a purely statistical density estimation task. The framework combines semantic feature abstraction, causal dependency modeling, and expert-guided reasoning to generate synthetic patient records that preserve clinical plausibility, interpretability, and privacy. An overview of the methodology is illustrated in Fig.1.

### A. Fuzzy Set Construction for Medical Data Presentation

Medical tabular data are characterized by heterogeneous feature types, measurement uncertainty, and semantically meaningful value ranges. To capture these properties, each feature is represented using fuzzy sets, which are linguistically interpretable and they are derived directly from the data and aligned with medical semantics. Let $M = \{(x_p, y_p)\}_{p=1}^{P}$, denote the training medical tabular dataset, where $x_p \in \mathbb{R}^D$ represents a patient record with $d$ medical variables and $y_p = 1, \ldots, G$ is the corresponding class label for a $G$ class.

Each feature $x_{p,d}, d = 1, \ldots, D$ is initially normalized to a common range [0,1] to ensure comparability across heterogeneous medical measurements.

The next step of the framework involves constructing appropriate fuzzy sets. Given a normalized medical dataset, each set of feature $M_g = \{(x_p, y_p) | y_p = g\}_{p=1}^{P}$ is analyzed independently for each class $G$ to identify representative value regions using clustering, usually fuzzy c-means [12]. Let $K^g = \{k_e^g\}_{e=1}^{E_g}$ denote the resulting cluster centers, where $k_e^g \in \mathbb{R}^D$ is the value of the center of the cluster $e = 1, \ldots, E_g$, formed by the features of the patient subset $M_g$. For each component of the cluster centers $k^g_{e,d}$ a corresponding fuzzy set $\tilde{A}^g_{e,d}$ is defined representing a linguistic qualifier associated with the $e$-th cluster and the $d$-th feature. This set is characterized by a membership function $\mu^g_{e,d}(x_{p,d}) \in [0,1]$ which for simplicity is considered triangular. The top of this triangular function is positioned at $k^g_{e,d}$ while its support extends to the nearest neighboring component $k^g_{e',d}$ , $e \neq e'$ or to the boundary values 0 or 1 using trapezoidal membership function, depending on which is closest. The membership functions provide a soft association between numerical feature values and linguistically expressed states, enabling gradual transitions and robust handling of borderline cases that commonly occur in medical measurements. In borderline cases e.g., between normal and high values, fuzzy membership provides a gradual transition rather than a hard assignment, which improves robustness and reflects clinical uncertainty. The region of each fuzzy set $\tilde{A}^g_{e,d}$ , for $e = 1, \ldots, E_g$ , corresponds to a fuzzy region of probable membership to clusters, which in turn represent medically meaningful states that can be semantically interpreted, e.g., low, normal, high values of feature $x_{p,d}$, capturing meaningful value ranges of each medical feature. The fuzzy sets $\tilde{A}^g_{e,d}$ correspond to the linguistic concept nodes of the FCM, linking the feature-level representation to the concept activations defined in Section II.

Although clustering identifies representative value regions, it does not indicate how clinically meaningful or prevalent each linguistic qualifier is within a patient group. To quantify this, the relevance of each linguistic state is estimated based on the degree to which patient samples of class $g$ activate the corresponding fuzzy set. For each fuzzy set $\tilde{A}^g_{e,d}$ the fuzzy cardinality [13] is computed by aggregating membership degrees across all samples for the specific feature dimension. This quantity reflects the cumulative compatibility between observed medical measurements of class $g$ and the linguistic concept represented by $\tilde{A}^g_{e,d}$. Precisely, it provides an estimation of how strongly a particular medical state, such as *High* blood pressure, is expressed within the patient data. The fuzzy cardinality therefore serves as an indicator of the activity level and representativeness of each linguistic concept within the class. The fuzzy sets are ordered in ascending order based on their cardinality. This ordering allows the framework to differentiate between dominant and marginal medical states. Concepts that belong to a fuzzy set with linguistic *High*, they can exert greater influence during subsequent causal reasoning. In this way, the representation remains aligned with the actual clinical profile of the patient population while preserving interpretability for expert inspection. In this context, the term linguistic does not imply predefined clinical labels but refers to interpretable fuzzy abstractions of feature values derived from data. The fuzzy sets represent relative states, e.g., lower, intermediate, higher values, rather than fixed medical categories. While clustering is data-driven and may not always align exactly with established clinical thresholds, the resulting fuzzy representation remains interpretable as an ordered and continuous description of feature intensity. Furthermore, the framework supports expert-driven refinement, allowing clinicians to relabel, merge, or adjust fuzzy sets to better align with medical semantics when required.

An illustrative example of the fuzzy set construction is shown in Fig. 2. In this example, E=2 clusters are identified for patients of class $g$. The horizontal axis represents normalized medical values, while $k^g_{1,1}, k^g_{2,2}$ denote the corresponding cluster centers for two different features $d = 1,2$. Due to normalization, different feature dimensions can be compared within a unified scale for interaction analysis. The fuzzy sets $\tilde{A}^g_{1,1}$ , $\tilde{A}^g_{2,2}$ define two linguistic states e.g., moderate and high response, and the vertical axis represents the membership degree $\mu^g_{e,d}(x_{p,d})$. From a medical decision support perspective, this representation transforms a raw numeric measurement into interpretable clinical states that can be directly incorporated into causal reasoning. For example, a patient has a condition where the features $x_{p,d}, d = 1,2$ belong one to moderate and the other to high response states, reflecting their causal relation. This concept-level abstraction supports transparent reasoning and enables clinicians to understand how each medical feature contributes to synthetic case generation and downstream decision support modeling. Although membership functions are defined per feature dimension, normalization of all features to [0,1] enables their representation in a shared domain. This allows the visualization and computation of cross-feature overlaps, which capture co-activation patterns between different features and are used to estimate inter-feature relationships in the FCM. It should be noted that the overlap shown in Fig. 2 does not imply that membership functions are jointly defined across features, rather, it provides an illustrative projection of independently defined feature-wise fuzzy sets to visualize their potential co-activation. The triangular overlap shown in Fig.2 is illustrative, in practice, overlap regions may take different shapes depending on the relative positions of the membership functions.

### B. Definition of FCM Interconnections

In the proposed framework, directed edges of the FCM represent clinically interpretable influence relationships between medical features. While such relationships are often defined directly by medical experts, in this study the initial edge weights are estimated from data using interactions between the corresponding fuzzy feature representations. In the FCM graph, the nodes represent the various linguistic characterizations associated with each feature $x_{p,d}$. Let $x_{p,i}$, $x_{p,j}$ denote two medical features for patient $p$ of class $g$. Each feature activates the nodes $C_i$ , $C_j$ of FCM, which are describe by the fuzzy sets $\tilde{A}^g_{e,i}$ , $\tilde{A}^g_{e',j}$ where $e \neq e'$ .The influence between these features is defined as the intersection of the fuzzy sets they belong as:

$$\tilde{B}_{ji} = \tilde{A}^g_{e,i} \cap \tilde{A}^g_{e',j} \tag{3}$$

which captures the co-activation strength of the two medical concepts (overlapping region in Fig.2). The directed weight $w_{ji}$ for concept nodes $C_i$ , $C_j$ is computed based on fuzzy center-of-gravity (CoG) of each fuzzy set $\tilde{B}_{ji}$. The resulting weights are normalized to the interval $[-1,1]$.

The FCM serves as an interpretable intermediate model that clinicians can review and adjust. Experts may validate whether detected dependencies align with known physiological mechanisms, remove spurious associations, enforce medically established causal directions, or constrain the sign of specific edges. For example, known risk factors may be restricted to exert only positive influence on disease-related concepts. This human-in-the-loop validation step transforms the FCM from a purely data-driven association model into a medical causal representation. By enabling expert oversight and structural refinement, the framework supports transparency, and trust, which are essential requirements for medical decision support systems.

### C. Medical Data Generation

A FCM constructed as described in the previous sections, can generate synthetic samples by jointly sampling linguistic feature states and enforcing causal consistency through fuzzy cognitive propagation. For each synthetic sample $\tilde{x} \in R^D$ , generation begins by randomly initializing a linguistic activation vector sampled from a uniform distribution over [0,1] in the fuzzy concept space rather than sampling numeric feature values directly. An initial activation vector is defined as $A^0 = [\mu^g_{0,0}, \dots, \mu^g_{e,d}, \dots, \mu^g_{E,D}]$ . The activation vector is iteratively propagated through the FCM using the equation (1) until convergence. This reasoning process enforces directed feature dependencies. Random initial states are therefore transformed into causally coherent configurations consistent with learned medical feature interactions.

Once a stable state is reached, crisp numerical inference is performed via defuzzification. Convergence is determined by the difference between successive activation vectors, $\| A^{t+1} - A^t \| < \epsilon$, with $\epsilon = 10^{-3}$, a commonly used tolerance for stable numerical convergence. A maximum of 100 iterations is imposed to guarantee termination, which was found sufficient for convergence across all datasets. More specifically, for each feature $\tilde{x}_d$ the most strongly activated fuzzy set is identified following the Maximum Membership principle, and a crisp value is sampled from their support, biased by activation strength. Thus, a controlled randomness is introduced so that synthetic records cover a realistic range of clinically plausible values rather than collapsing to a fixed prototype. In cases where multiple fuzzy sets are simultaneously activated for a given feature, a weighted defuzzification strategy is applied, producing a crisp value as a weighted combination of the fuzzy set centers, with weights proportional to their respective activation strengths [14].

This generation process remains open to expert intervention, *i.e.*, medical experts are able to revise selected concept values, enforce known clinical rules, or adjust edge weights prior to propagation. During the data generation process, experts can inspect stabilized concept states and reject implausible configurations before defuzzification. This human-in-the-loop capability ensures that synthetic samples align with established medical knowledge.

## IV. Experiments and Results

### A. Experimental Setup

The experimental evaluation aims to assess the effectiveness of the proposed fuzzy concept-based framework with FCM for synthetic medical data generation and to compare its performance against established baseline methods under comparable computational conditions. All experiments were conducted using CPU-only execution, without GPU acceleration. This choice reflects the design objective of the proposed approach as a computationally efficient and accessible alternative to resource-intensive deep generative models. To ensure fairness and reproducibility, comparisons were limited to widely adopted tabular data synthesizers that can be reliably trained in CPU-based environments. Models requiring extensive GPU resources were excluded to maintain consistent experimental conditions.

The proposed method was evaluated against three representative synthesizers from the Synthetic Data Vault (SDV) framework [15]. Gaussian Copula, a classical statistical baseline, CTGAN, a GAN tailored for tabular data and TVAE, a VAE-based approach were utilized. All baseline models were implemented using their default SDV configurations. For each dataset, the number of generated synthetic samples was set equal to the size of the corresponding real training set to ensure comparable data availability across methods.

Experimental evaluation was conducted on three publicly available datasets from the UCI repository [16] : Pima Indians Diabetes (768 instances, 8 features; 500 without and 268 with diabetes), South African Heart Disease (462 instances, 9 features; 302 without and 160 with heart disease), and Statlog Heart Disease (270 instances, 13 features; 150 without and 120 with heart disease). These datasets contain mixed continuous and categorical features and are widely used in tabular classification tasks. For each dataset, a consistent train–test split of 80/20 was applied across all methods. Synthetic data were generated exclusively from the training portion, while evaluation metrics were computed using the real test set to assess generalization and downstream utility.

### B. Results

The evaluation followed a Train-on-Synthetic-Test-on-Real (TSTR) protocol, which is widely regarded as a reliable indicator of synthetic data utility. Under this protocol, predictive models are trained exclusively on synthetic data and evaluated on unseen real test data to assess generalization performance. Evaluation encompassed utility, fidelity, and privacy dimensions using a consistent pipeline across all methods[15].

TABLE I
Synthetic Medical Data Evaluation

| Dataset | **CTGAN** | **Gaussian Copula** | **TVAE** | **Proposed Method** |
|---|---|---|---|---|
| UTILITY METRICS | | | | |
| **Accuracy ↑*** | | | | |
| Diabetes | 0.66 | 0.71 | 0.72 | **0.73** |
| SAHeart | 0.60 | 0.63 | **0.73** | 0.72 |
| Heart Disease | 0.69 | 0.74 | 0.80 | **0.81** |
| **F1-score ↑** | | | | |
| Diabetes | 0.63 | 0.67 | **0.73** | 0.70 |
| SAHeart | 0.51 | 0.58 | **0.68** | 0.64 |
| Heart Disease | 0.63 | 0.74 | **0.82** | 0.78 |
| **AUROC ↑** | | | | |
| Diabetes | 0.64 | 0.71 | **0.80** | 0.77 |
| SAHeart | 0.48 | 0.62 | **0.73** | 0.70 |
| Heart Disease | 0.56 | 0.87 | 0.89 | **0.90** |
| FIDELITY METRICS | | | | |
| **KS Complement ↑** | | | | |
| Diabetes | 0.74 | 0.82 | **0.85** | 0.80 |
| SAHeart | 0.75 | **0.90** | 0.82 | **0.90** |
| Heart Disease | 0.85 | **0.93** | 0.85 | 0.91 |
| **Correlation Similarity ↑** | | | | |
| Diabetes | 0.90 | 0.93 | **0.95** | **0.95** |
| SAHeart | 0.88 | **0.95** | 0.93 | 0.89 |
| Heart Disease | 0.91 | **0.94** | **0.94** | **0.94** |
| **Statistical Similarity ↑** | | | | |
| Diabetes | 0.92 | 0.93 | **0.95** | **0.95** |
| SAHeart | 0.92 | **0.97** | 0.93 | 0.95 |
| Heart Disease | 0.92 | **0.98** | 0.90 | 0.93 |
| PRIVACY METRICS | | | | |
| **DCR Baseline Protection ↑** | | | | |
| Diabetes | **0.43** | 0.4 | 0.21 | 0.40 |
| SAHeart | **0.49** | 0.35 | 0.27 | 0.35 |
| Heart Disease | **0.24** | 0.18 | 0.07 | 0.20 |

* The arrows ↑ indicate that higher values represent better scores.

Utility was measured by training a classification model on synthetic data and evaluating it on real test data using Accuracy, F1-score, and Area Under the ROC Curve (AUROC). Fidelity assessed the preservation of statistical characteristics of the original data through Kolmogorov–Smirnov (KS) Complement for marginal distribution, reported as 1 minus the KS statistic, so that higher values indicate greater similarity, Correlation Similarity for inter-feature dependency preservation, and Statistical Similarity for agreement in aggregated moment-based statistics. Privacy risk was evaluated using Distance to Closest Record (DCR) Baseline Protection. For each synthetic record, DCR measures its distance to the nearest real training record higher values indicate that synthetic samples are farther from any real record, implying lower re-identification risk. Visual comparison of feature distributions can further support the assessment of similarity between real and synthetic data.

The results demonstrate that the proposed FCM-based framework achieves strong performance across all three evaluation dimensions (Table I). Under the TSTR setting, the method consistently outperforms CTGAN and remains competitive with TVAE and Gaussian Copula in terms of Accuracy, F1-score, and AUROC, while maintaining substantially lower computational complexity. In particular, on the Heart Disease dataset, AUROC values are nearly identical to those achieved by TVAE despite operating under strict CPU-only constraints and linear complexity. Fidelity analysis confirms effective preservation of marginal distributions and feature relationships, as reflected in high KS Complement and correlation similarity scores. Although Gaussian Copula achieves marginally higher fidelity in certain cases, this is often accompanied by reduced predictive utility, whereas the proposed framework maintains a balanced trade-off between realism and downstream task performance.

Privacy evaluation indicates moderate-to-strong DCR baseline protection, consistently exceeding that of TVAE, where higher scores reflect greater distance between synthetic and real records, indicating lower re-identification risk. Unlike deep generative models that depend on computationally intensive neural optimization, the proposed approach leverages sparse linguistic activations and causal propagation within an FCM structure. Although the theoretical size of the FCM scales with the number of features and linguistic states, only a limited subset of concepts becomes active during reasoning, resulting in efficient inference. Overall, these findings indicate that high-quality synthetic tabular data can be generated through interpretable, causally grounded fuzzy modeling while operating under strict computational constraints, making the approach suitable for resource-aware and transparency-critical clinical decision support applications.

## V. Conclusion

This paper introduced a fuzzy concept-based framework for synthetic tabular data generation that models the generation process as causally guided fuzzy reasoning rather than conventional density estimation. Feature marginals are represented through fuzzy sets, while inter-feature dependencies are captured using sparse Fuzzy Cognitive Maps (FCMs), enabling interpretable and computationally efficient generation.

Synthetic records are produced by transforming randomly initialized linguistic profiles into causally stabilized concept states and subsequently into numerical feature vectors. This mechanism promotes structural coherence, semantic transparency, and expert oversight, which are essential properties for clinical decision support applications. Experimental results across multiple benchmark datasets demonstrate competitive utility, fidelity, and privacy compared to widely used baselines, while operating under strict CPU-only constraints and avoiding the complexity of deep generative models. The findings indicate that causally grounded fuzzy modeling provides a practical and resource-efficient alternative for synthetic medical tabular data generation. Future work includes automated structure refinement and extensions to temporal medical data, and the incorporation of explicit multivariate similarity metrics to better assess the preservation of joint feature distributions in synthetic data.

## References

[1] A. Kurakova and H. Homayouni, “A comprehensive evaluation framework for synthetic medical tabular data generation,” *Journal of Biomedical Informatics*, p. 104939, 2025.

[2] G. C. N. Kindji, L. M. Rojas-Barahona, E. Fromont, and T. Urvoy, “Tabular data generation models: An in-depth survey and performance benchmarks with extensive tuning,” *Neurocomputing*, p. 131655, 2025.

[3] L. Xu, M. Skoularidou, A. Cuesta-Infante, and K. Veeramachaneni, “Modeling tabular data using conditional gan,” *Advances in neural information processing systems*, vol. 32, 2019.

[4] R. B. Nelsen, *An introduction to copulas*. Springer, 2006.

[5] P. A. Apellaniz, B. A. Galende, A. Jim, J. Parras, S. Zazo, and others, “Advancing Cancer Research with Synthetic Data Generation in Low-Data Scenarios,” *IEEE Journal of Biomedical and Health Informatics*, 2025.

[6] C. Umesh, K. Schultz, M. Mahendra, S. Bej, and O. Wolkenhauer, “Preserving logical and functional dependencies in synthetic tabular data,” *Pattern Recognition*, vol. 163, p. 111459, 2025.

[7] M. D. Vasilakakis and D. K. Iakovidis, “Fuzzy similarity phrases for interpretable data classification,” *Information Sciences*, vol. 624, pp. 881–907, 2023.

[8] B. Kosko, “Fuzzy cognitive maps,” *International journal of man-machine studies*, vol. 24, no. 1, pp. 65–75, 1986.

[9] M. D. Vasilakakis, G. Triantafyllou, and D. K. Iakovidis, “Intuitionistic Fuzzy Cognitive Map for Disease Transmission Risk Assessment in Ventilated Indoor Spaces,” *IEEE Access*, vol. 13, pp. 201789–201801, 2025.

[10] Y. Boutalis, T. L. Kottas, and M. Christodoulou, “Adaptive estimation of fuzzy cognitive maps with proven stability and parameter convergence,” *IEEE Transactions on Fuzzy Systems*, vol. 17, no. 4, pp. 874–889, 2009.

[11] G. Sovatzidi, M. D. Vasilakakis, and D. K. Iakovidis, “Automatic fuzzy graph construction for interpretable image classification,” in *2022 IEEE International Conference on Image Processing (ICIP)*, 2022, pp. 3743–3747.

[12] J. C. Bezdek, *Pattern recognition with fuzzy objective function algorithms*. Springer Science & Business Media, 2013.

[13] L. A. Zadeh, “A computational approach to fuzzy quantifiers in natural languages,” in *Computational linguistics*, Elsevier, 1983, pp. 149–184.

[14] W. Van Leekwijck and E. E. Kerre, “Defuzzification: criteria and classification,” *Fuzzy sets and systems*, vol. 108, no. 2, pp. 159–178, 1999.

[15] N. Patki, R. Wedge, and K. Veeramachaneni, “The synthetic data vault,” in *2016 IEEE international conference on data science and advanced analytics (DSAA)*, 2016, pp. 399–410.

[16] C. J. Merz, “UCI repository of machine learning databases,” *URL: http://www. ics. uci. edu/ mlearn/MLRepository. html*, 1998.